%% file: ms.tex
\documentclass[10pt,twocolumn,letterpaper]{article}
\pdfoutput=1
\usepackage{cvpr}
\usepackage{times}
\usepackage{epsfig}
\usepackage{threeparttable}
\usepackage{graphicx}
\usepackage{amsmath}
\usepackage{textcomp}
\usepackage{hyperref}
\usepackage{import}
\usepackage{graphics}
\usepackage{booktabs}
\usepackage{caption}
\usepackage{subcaption}
\usepackage{enumitem}
\usepackage{tabularx}
\usepackage{multirow}
\usepackage{stfloats}
\usepackage{amssymb}
\usepackage{latexsym}
\usepackage[title]{appendix}
\usepackage{authblk}

\cvprfinalcopy 

\def\cvprPaperID{5508} 

\newcommand\numberthis[1][]{%
    \refstepcounter{equation}%
    \ifx#1\empty\else\label{eq:#1}\fi%
    \tag{\theequation}%
}

\begin{document}

\title{Good News, Everyone! Context driven entity-aware captioning for news images}








\author{Ali Furkan Biten, 
Lluis Gomez,
Mar\c{c}al Rusi{\~n}ol, 
Dimosthenis Karatzas\\
Computer Vision Center, UAB, Spain\\
{\tt\small \{abiten, lgomez, marcal, dimos\}@cvc.uab.es}
}

\maketitle

\begin{abstract}
Current image captioning systems perform at a merely descriptive level, essentially enumerating the objects in the scene and their relations. Humans, on the contrary, interpret images by integrating several sources of prior knowledge of the world.
In this work, we aim to take a step closer to producing captions that offer a plausible interpretation of the scene, by integrating such contextual information into the captioning pipeline. For this we focus on the captioning of images used to illustrate news articles. We propose a novel captioning method that is able to leverage contextual information provided by the text of news articles associated with an image. Our model is able to selectively draw information from the article guided by visual cues, and to dynamically extend the output dictionary to out-of-vocabulary named entities that appear in the context source. Furthermore we introduce ``GoodNews'', the largest news image captioning dataset in the literature and demonstrate state-of-the-art results.
\end{abstract}

\section{Introduction}
People understand scenes by building causal models and employing them to compose stories that explain their perceptual observations~\cite{lake2017building}. This capacity of humans is associated with intelligent behaviour. One of the cognitive tasks in the Binet-Simon intelligence test~\cite{terman1916measurement} is to describe an image.
Three performance levels are defined, going from enumeration of objects in the scene, to basic description of contents and finally interpretation, where contextual information is drawn upon to compose an explanation of the depicted events.
%
\begin{figure}[htp]
\begin{center}
\includegraphics[width=\linewidth,height=0.3\textwidth]{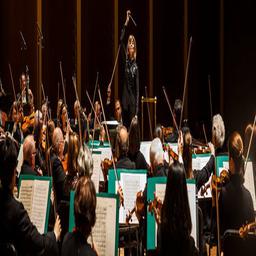} 
\setlength\extrarowheight{3pt} 
\begin{tabularx}{\linewidth}{@{\extracolsep{\fill}} |X|}
\small{{\textbf{Ground Truth}: JoAnn Falletta leading a performance of the Buffalo Philharmonic Orchestra at Kleinhans Music Hall.}} \\
\small{{\textbf{Show \& Tell} \cite{vinyals2015show}: A group of people standing around a table.}} \\
\small{{\textbf{Ours}: JoAnn Falletta performing at the Buffalo Philharmonic Orchestra.}} \\
\hline
\end{tabularx}
\caption{Standard approaches to image captioning cannot properly take any contextual information into account. Our model is capable of producing captions that include out-of-vocabulary named entities by leveraging information from available context knowledge.}
\label{main}
\end{center}
\vspace{-1.0cm}
\end{figure}

Current image captioning systems \cite{vinyals2015show, anderson2017bottom, Karpathy, rennie2017self, lu2017knowing, Fang2015} can at best perform at the description level, if not restricted at the enumeration part, while failing to integrate any prior world knowledge in the produced caption.
Prior world knowledge might come in the form of social, political, geographic or temporal context, behavioural cues, or previously built knowledge about entities such as people, places or landmarks.
In this work, we aim to take a step closer to producing captions that offer a plausible interpretation of the scene, by integrating such contextual information into the captioning pipeline. 

This introduces numerous new challenges.
On one hand, the context source needs to be encoded and information selectively drawn from it, guided by the visual scene content.
On the other hand, explicit contextual information, typically found in the form of named entities such as proper names, prices, locations, dates, etc, which are typically out-of-dictionary terms or at best underrepresented in the statistics of the dictionary used, need to be properly injected in the produced natural language output.

Currently available image captioning datasets are not fit for developing captioning models with the aforementioned characteristics, as they provide generic, dry, repetitive and non-contextualized captions, 
while at the same time there is no contextual information available for each image. For the task at hand, we considered instead other image sources, such as historical archive images or images illustrating newspaper articles, for which captions (i.e. descriptions provided by archivists, captions provided by journalists) and certain contextual information (i.e. history texts, news articles) is readily available or can be collected with reasonable effort. 

In this work, we focus on the captioning of images used to illustrate news articles. Newspapers are an excellent domain for moving towards human-like captions, as they provide readily available contextual information that can be modelled and exploited. In this case contextual information is provided by the text of the associated news article, along with other metadata such as titles and keywords. At the same time, there is readily available ground truth in the form of the existing caption written by domain experts (journalists), which is invaluable in itself. Finally, data is freely available at a large scale online. To this end, we have put together “GoodNews” the biggest news-captioning dataset in the literature with more than 466,000 images and their respective captions and associated articles.

To the best of our knowledge, generative news image captioning has been scarcely explored in the literature~\cite{feng2013automatic, tariq2017context,ramisa2018breakingnews}. Similarly to~\cite{ramisa2018breakingnews} we draw contextual information about the image from the associated article. Unlike ~\cite{ramisa2018breakingnews} which uses world-level encoding, we encode the article at the sentence level, as semantic similarity is easier to establish at this granularity. In addition, we introduce an attention mechanism in order to selectively draw information from the article guided by the visual content of the image.

News articles and their respective news image captions, unlike common image captioning datasets such as MSCOCO \cite{lin2014microsoft}, or Flickr \cite{plummer2015flickr30k}, contain a significant amount of named entities. Named entities\footnote{Named entities are the objects that can be denoted with a proper name such as persons, organizations, places, dates, percentages, etc. \cite{nadeau2007survey}} pose serious problems to current captioning systems that have no mechanism to deal with out-of-vocabulary (OOV) words. This includes~\cite{ramisa2018breakingnews} where named entity usage is implicitly restricted to the ones that appear in adequate statistics in the training set. Unlike existing approaches, we propose here an end-to-end, two-stage process, where first template captions are produced in which named entities’ placeholders are indicated along with their respective tags. These are subsequently substituted by selecting the best matching entities from the article, allowing our model to produce captions that include out-of-vocabulary words.

The contributions of this work are as follows:
\begin{itemize}[noitemsep,topsep=0pt]
    \item We propose a novel captioning method, able to leverage contextual information to produce image captions at the scene interpretation level. 
   \item We propose a two-stage, end-to-end architecture, that allows us to dynamically extend the output dictionary to out-of-vocabulary named entities that appear in the context source.
    \item We introduce “GoodNews”, the largest news image captioning dataset in the literature, comprising 466,000 image-caption pairs, along with metadata.
\end{itemize}

We compare the performance of our proposed method against existing methods and demonstrate state-of-the-art results. Comparative studies demonstrate the importance of properly treating named entities, and the benefits of considering contextual information. Finally, comparisons against human performance highlight the difficulty of the task and limitations of current evaluation metrics.

\section{Related Work}
\label{related}
Automatic image captioning has received increased attention lately as a result of advances in both computer vision and natural language processing stemming from deep learning~\cite{bai2018survey, bernardi2016automatic}.
Latest state-of-the-art models~\cite{Xu2015, lu2017knowing,rennie2017self, anderson2017bottom} usually follow an attention guided encoder-decoder strategy, in which visual information is extracted from images by deep CNNs and then natural language descriptions are generated with RNNs. Despite the good results current state-of-the-art models start to yield according to standard performance evaluation metrics, automatic image captioning is still a challenging problem.
Present-day methods tend to produce repetitive, simple 
sentences~\cite{devlin2015exploring} written in a consistent style, generally limited on enumerating or describing visual contents, and not offering any deeper semantic interpretation.

The latest attempts of producing richer human-like sentences, are centered in gathering new datasets that might be representative of different writing styles. For example, using crowd-sourcing tools to collect different styles of captions (negative/positive, romantic, humorous, etc.) as in~\cite{mathews2016senticap, gan2017stylenet}, or leveraging the usage of romance novels to change the style of captions to story-like sentences like in~\cite{mathews2018semstyle}. Even though gathering annotations with heterogeneous styles helps mitigating the repetitiveness of the outputs' tone, content-wise captions remain detailed descriptions of the visual content. Automatic captioning still 
suffers from a huge semantic gap referring to the lack of correlation between images and semantic concepts~\cite{tariq2017context}.


The particular domain of news image captioning, has been explored in the past towards incorporating contextual information to the produced captions.
In~\cite{feng2013automatic} 3K news articles were gathered from BBC News. Image captions were then produced by either choosing the closest sentence in the article or using a template-based linguistic method. In~\cite{tariq2017context}, 100K images were collected from TIME magazine, and refined the captioning strategy proposed by Feng et. al.~\cite{feng2013automatic}.

Closer to our work, Ramisa et. al.~\cite{ramisa2018breakingnews} (BreakingNews) used pre-trained word2vec representations of the news articles concatenated with CNN visual features to feed the generative LSTM.
A clear indicator of whether contextual information is correctly incorporated in such cases, is to check to what extent the produced image captions include the correct named entities given the context. This is a challenging task, as in most of the cases such named entities are only becoming available at test time.
Although this is particularly important in the case of news image captioning, to the best of our knowledge none of the existing methods addresses named entity inclusion, employing instead closed dictionaries.

Nevertheless, the problem of dealing with named entities has been explored in generic (not context-driven) image captioning. In~\cite{tran2016rich} after gathering Instagram data, a CNN is used to recognize celebrities and landmarks as well as visual concepts such as water, mountain, boat, etc. Afterwards, a confidence model is used to choose whether or not to produce captions with proper names or with visual concepts. In~\cite{lu2018entity} template captions were created using named entity tags, that were later filled by the usage of a knowledge-base graph. The aforementioned methods require a predefined set of named entities. Unlike these methods, our approach looks in the text while producing a caption and ``attends'' to different sentences for entity extraction, which makes our model consider the context in which the named entities appear
to incorporate new, out-of-vocabulary named entities in the produced captions.

\section{The GoodNews Dataset}
\label{dataset}

To assemble the \textit{GoodNews} dataset, we have used the New York Times API to retrieve the URLs of news articles ranging from 2010 to 2018. We will provide the URLs of the articles and the script to download images and related metadata, also the released scripts can be used to obtain $167$ years worth of news. However, for image captioning purposes, we have restricted our collection to the last $8$ years of data, mainly because it covers a period when images were widely used to illustrate news articles. In total, we have gathered $466,000$ images with captions, headlines and text articles, randomly split into $424,000$ for training, $18,000$ for validation and $23,000$ for testing. 

\textit{GoodNews} exhibits important differences to current benchmark datasets for generic captioning like MSCOCO, while it is similar in nature, but about five times larger than BreakingNews, the largest currently available dataset for news image captioning. Key aspects are summarized in~\autoref{comparison}. The \textit{GoodNews} dataset, similarly to BreakingNews, exhibits longer average caption lengths than generic captioning datasets like MSCOCO, indicating that news captions tend to be more descriptive. 

\textit{GoodNews} only includes a single ground truth caption per image, while MSCOCO offers $5$ different ground truth captions per image. However, \textit{GoodNews} captions were written by expert journalists, instead of being crowd-sourced, which has implications to the style and richness of the text.
\begin{table}[ht!]
\scriptsize
\caption{Comparison of captioning datasets.}
\begin{tabular}{lrrrr}
\toprule
                   & {MSCOCO}  & {{BreakingNews}} & {{GoodNews}}\\
                  \midrule
Number of Samples          & 120k         &    110k & 466k                   \\
Average Caption Length (words)          & 11.30               &    28.09  &    18.21         \\
Named Entities{(Word)}  & 0\%          &     15.66\%  & 19.59\%       \\
Named Entities{ (Sentence)}  & 0\% &   90.79\%  &   95.51\%              \\
Nouns              & 33.45\%& 55.59\%         & 46.70\%      \\
Adjectives         & 27.23\%& 7.21\%        & 5\%     \\
Verbs              & 10.72\% & 12.57\%       & 11.22\%    \\
Pronouns           & 1.23\%& 1.36\%      & 2.54\%          \\ \bottomrule
\end{tabular}
\label{comparison}
\end{table}

Named entities represent $20\%$ of the words in the captions of \textit{GoodNews}, while named entities are by design completely absent from the captions of MSCOCO.
At the level of sentences, $95\%$ of caption sentences and $73\%$ of article sentences in \textit{GoodNews} contain at least one named entity. Moreover, we observe that \textit{GoodNews} has more named entities than BreakingNews at both token level and sentence level. 
Analyzing the part of speech tags, we observe that both \textit{GoodNews} and BreakingNews have less amount of adjectives but a higher amount of verbs and significantly higher amount of pronouns and nouns than MSCOCO. Given the nature of news image captions, this is expected, since they do not describe scene objects, but rather offer a contextualized interpretation of the scene. 

A key difference between our dataset and BreakingNews, apart from the fact that \textit{GoodNews} has five times more samples, is that our dataset includes a wider range of events and stories since \textit{GoodNews} spans a much longer time period. On the other hand, we must point out that BreakingNews offers a wider range of metadata as it aims to more tasks than news image captioning.

\section{Model}
\label{model}
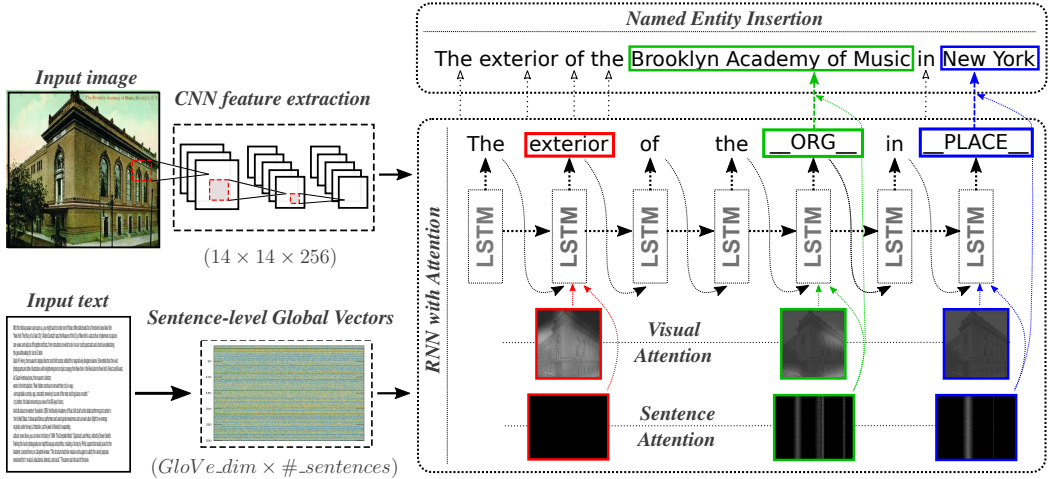
\begin{figure*}
\centering
\resizebox{0.8\linewidth}{!}{\import{./}{model_figure.tex}} 
\caption{Overview of our model where we combine the visual and textual features to generate first the template captions. Afterwards, we fill these templates with the attention values obtained over the input text. (Best viewed in color)}
\label{fig:model}
\end{figure*}

As illustrated in \autoref{fig:model} our model for context driven entity-aware captioning consists of two consecutive stages. In the first stage, given an image and the text of the corresponding news article, our model generates a template caption where placeholders are introduced to indicate the positions of named entities. In a subsequent stage our model selects the right named entities to fill those placeholders with the help of an attention mechanism over the text of the news article. 

We have used SpaCy's named entity recognizer~\cite{spacy2} to recognize named entities in both captions and articles of the \textit{GoodNews} dataset. We create template captions by replacing the named entities with their respective tags. At the article level, we store the named entities to be used later in the named entity insertion stage (see \autoref{insertion}). As an example, the caption ``Albert Einstein taught in Princeton University in 1921'' is converted into the following template caption: ``PERSON\_ taught in ORGANIZATION\_ in DATE\_''. The template captions created this way comprise the training set ground truth we use to train our models.
Our model is designed as a two-stream architecture, that combines a visual input (the image) and a textual input (the encoding of the news article).

Our model's main novelty comes from the fact that it encodes the text article associated with each input image and uses it as a second input stream, while employing an attention mechanism over the textual features. For encoding the input text articles we have used the Global Vectors (GloVe) word embedding~\cite{pennington2014glove} and an aggregation technique to obtain the article sentence level features. The attention mechanism provides our model with the ability to focus on distinct parts (sentences) of the article at each timestep. Besides, it makes our model end-to-end, capable of inserting the correct named entity in the template caption at each timestep using attention, see~\autoref{fig:model}. 


\subsection{Template Caption Generation}
\label{articleEnc}
For the template caption generation stage we follow the same formulation as in state-of-the-art captioning systems~\cite{Xu2015, lu2017knowing, anderson2017bottom} which is to produce a word at each timestep given the previously produced word and the attended image features in each step, trained with cross entropy. More formally, we produce a sentence $s_i := \{ w_0, w_1, ..., w_N\}$, where  $w_i$ is a one-hot vector for the $i$th word, as follows:

\allowdisplaybreaks
\begin{gather*}
x_t = W_e * w_t , \ where \ t \in \ {\{0, 1, ..., N-1\}}, \\
o_t = LSTM( concat(x_t, I_t, A_t) ), \\
w_{t+1} = softmax(W_{ie}o_t),  \\
L = - \sum_{t=0}^{N} log(w_{t+1}) \numberthis
\end{gather*}
where $W_e$, $W_{ie}$ are learnable parameters, $A_t$ denotes attended article features, and $I_t$ the attended image features. 
The attended image features at timestep $t$ are obtained as a function of the hidden state of previous timestep and the image features extracted using a Deep CNN model:
\begin{equation}
    \begin{split}
    &
I_f = CNN(I),\\&
I_t = Att(h_{t-1}, I_f)
    \end{split}
\end{equation}
where $h_{t-1}$ is the hidden state at time $t-1$, $I$ is the input image, and $I_f$ are features of the input image extracted from a ResNet \cite{he2016deep} network pretrained on ImageNet \cite{russakovsky2015imagenet}.

In the next section we describe three different article encoding techniques that we have used to obtain a fixed size matrix $A_f$ with the sentence level features of the input article. Later, we will explain in detail how we  calculate the attended article features, $A_t$, at every timestep $t$. 

\subsection{Article Encoding Methods} 
Inspired by the state of the art on semantic textual similarity tasks~\cite{arora2016simple}, we use a sentence level encoding to represent the news articles in our model, as domain, purpose and context are better preserved at the sentence level.

By using a sentence level encoding, we overcome two shortcomings associated with word level encodings. First, encoding the article at the word granularity requires a higher dimensional matrix which makes the models slower to train and converge. Second, a word level encoding cannot encode the flow (or context) that sentences provide, e.g. ``He graduated from Massachusetts'' and ``He is from Massachusetts'': the former is for MIT which is an organization while the latter one is a state. 

Formally, to obtain the sentence level features for the $i^{th}$ article, $A_i := \{s^{art}_0, s^{art}_1, ..., s^{art}_M\}$, where $s^{art}_j = \{w_0, w_1, ..., w_{N_j} \}$ is the $j^{th}$ sentence of article and $w_k$ is the word vector obtained from the pre-trained GloVe model, we have first used a simple average of words for each sentence of the article:
\begin{equation}
\begin{split}
    A_{f_j}^{avg} = \dfrac{1}{N_j} \sum_{i=0}^{N_j} w_i, where \ j = {0, 1, ..., M}
\end{split}
\end{equation}
As an alternative we have also considered the use of a weighted average of word vectors according to their smoothed inverse frequency because the simple average of word vectors has huge components along semantically meaningless directions~\cite{arora2016simple}:
\begin{equation}
\begin{split}
A_{f_j}^{wAvg} = \dfrac{1}{N_j} \sum_{i=0}^{N_j} p(w_i) * w_i, \ p(w) =\frac{a}{a + tf(w)}
\end{split}
\end{equation}

Finally, we have explored the use of the tough-to-beat baseline (TBB)~\cite{arora2016simple}, which consists in subtracting the first component of the PCA from the weighted average of the article encoding since empirically the top singular vectors of the datasets seem to roughly correspond to the syntactic information or common words:
\begin{equation}
\begin{split}
&
A_{f_j}^{wAvg} = U \ \Gamma \ V, \\&
X = U^* \ \Gamma^* \ V^*,\ where\ X\ is\ the\ 1^{st}\ component \\&
A_{f_j}^{TBB} = A_f^{wAvg} - X
\end{split}
\end{equation}

\textbf{Article Encoding with Attention:}
After obtaining the article sentence level features, $A_f \in R^{M \times D_w}$, where $M$ is the fixed sentence length and $D_w$ is the dimension of the word embedding, we have designed an attention mechanism that works by multiplying the sentence level features with an attention vector $\beta_t \in R^M$ :
\begin{equation}
\begin{split}
&
    A_f = GloVe(A_i), \\ &
    A_t = \beta_t * A_f  
\end{split}
\end{equation}
where given the previous timestep of the LSTM, $h_{t-1}$ and article features, $A_f$, we learn the attention with a fully connected layer:
\begin{equation}
\begin{split}
&
    \theta_t = FC(h_{t-1}, A_f), \\&
    \beta_t = softmax(\theta_t)
\end{split}
\end{equation}

As explained next, apart from improving the generation of the template captions, the usage of attention enables us to also to select the correct named entities to include on the basis of the attention vector.
\subsection{Named Entity Insertion}
\label{insertion}
After generating the template captions, we insert named entities according to their categories. If there are more than one tag of PERSON, ORGANIZATION, LOCATION, etc. in the top ranked sentence, we select the named entity in order of appearance in the sentence. In order to compare our method with standard image captioning models we came up with there different insertion techniques, from which two can be used with visual-only architectures (i.e. without considering the article text features):  Random Insertion (RandIns) and Context-based Insertion (CtxIns). Whereas the third one is based on an attention mechanism over the article that guides the insertion (AttIns).

The random insertion (RandIns) offers a baseline for the other insertion methods explored, and it consists of randomly picking a named entity of the same category from the article, for each named entity placeholder that is produced in the template captions. 

For the Context Insertion (CtxIns) we make use of a pretrained GloVe embedding to rank the sentences of articles with cosine similarity 
according to the produced template caption embedding
and afterwards insert the named entities on the basis of this ranking. 

Finally, for our insertion by attention method (AttIns), we use the article attention vector $\beta_t$ that is produced at each timestep $t$ of the template caption generation to insert named entities without using any external insertion method. 

\begin{table*}[tp]
\centering
\caption{Results on the intermediate task of template caption generation for state-of-the-art captioning models without using any Article Encoding (top) and for our method using different Article Encoding strategies (bottom).}
\scriptsize
\begin{tabular}{lllllllll}
\toprule
                 & Bleu-1   & Bleu-2  & Bleu-3  & Bleu-4  & Meteor   & Rouge-L    & CIDEr    & Spice    \\ \midrule

Show Attend Tell \cite{Xu2015} &\textbf{  11.537}\% & \textbf{5.757}\% & \textbf{2.983}\% &\textbf{1.711}\% & \textbf{13.559}\% & \textbf{20.468}\% & \textbf{17.317}\% & 22.864\% \\
Att2in2  \cite{rennie2017self}        & 10.536\% & 5.176\% & 2.716\% & 1.542\% & 12.962\% & 19.934\% & 16.511\% & 23.789\% \\
Up-Down  \cite{anderson2017bottom}        & 10.812\% & 5.201\% & 2.649\% & 1.463\% & 12.546\% & 19.424\% & 15.345\% & 23.112\% \\
Adaptive Att \cite{lu2017knowing}    & 7.916\%  & 3.858\% & 1.941\% & 1.083\% & 12.576\% & 19.638\% & 15.928\% & \textbf{25.017}\% \\
\midrule
Ours (Average)          & \textbf{13.419}\% & \textbf{6.530}\% & \textbf{3.336}\% & \textbf{1.869}\% & \textbf{13.752}\% & \textbf{20.468}\% & \textbf{17.577}\% & 22.699\% \\
Ours (Weighted Average) & 11.898\% & 5.857\% & 3.012\% & 1.695\% & 13.645\% & 20.355\% & 17.132\% & \textbf{23.251}\% \\
Ours (TBB)              & 12.236\% & 5.817\% & 2.950\% & 1.662\% & 13.530\% & 20.353\% & 16.624\% & 22.766\% \\ \bottomrule
\end{tabular}
\label{FirstNoArticle}
\label{FirstArticle}
\end{table*}

\begin{table*}[t]
\centering

\caption{Results on news image captioning. RandIns: Random Insertion; CtxIns: GloVe Insertion; AttIns: Insertion by Attention; No-NE: without named entity insertion.}
\begin{threeparttable}
\scriptsize
\begin{tabular}{clllllllll}
\toprule
&   & Bleu-1  & Bleu-2  & Bleu-3  & Bleu-4  & Meteor  & Rouge    & CIDEr    & Spice \\
\midrule
\parbox[t]{2mm}{\multirow{9}{*}{\rotatebox[origin=c]{90}{\textbf{Visual only}}}} & \textbf{Show Attend Tell - No-NE}  & 8.80\% & 3.01\% & 0.97\% & 0.43\% & 2.47\% & 9.06\% & 1.67\% & 0.69\% \\

& \textbf{Show Attend Tell} + RandIns & 7.37\% & 2.94\% & 1.34\% & 0.70\% & 3.77\% & 11.15\% & 10.03\% & 3.48\%      \\
&\textbf{Att2in2}  + RandIns         & 6.88\% & 2.82\% & 1.35\% & 0.73\% & 3.57\% & 10.84\% & 9.68\%  & 3.57\%      \\
&\textbf{Up-Down}  + RandIns & 6.92\% & 2.77\% & 1.29\% & 0.67\% & 3.40\% & 10.38\% & 8.94\%  & 3.60\%\\
&\textbf{Adaptive Att}  + RandIns & 5.22\% & 2.11\% & 0.97\% & 0.51\% & 3.28\% & 10.21\% & 8.68\%  & 3.56\%  \\

&\textbf{Show Attend Tell}  + CtxIns & 7.63\% & 3.03\% & 1.39\% & 0.73\% & 4.14\% & 11.88\% & 12.15\% &  4.03\%     \\
&\textbf{Att2in2}  + CtxIns         & 7.11\% & 2.91\% & 1.39\% & 0.76\% & 3.90\% & 11.58\% & 11.58\% & 4.12\%\\
&\textbf{Up-Down}  + CtxIns         & 7.21\% & 2.87\% & 1.34\% & 0.71\% & 3.74\% & 11.06\% & 11.02\% & 3.91\%\\
&\textbf{Adaptive Att}  + CtxIns    & 5.30\% & 2.11\% & 0.98\% & 0.51\% & 3.59\% & 10.94\% & 10.55\% & 4.13\%\\

\midrule
\parbox[t]{2mm}{\multirow{6}{*}{\rotatebox[origin=c]{90}{
\textbf{Visual \& Textual}}}} &  \textbf{BreakingNews*} - No-NE \cite{ramisa2018breakingnews} & 5.06\% & 1.70\% & 0.60\% &0.31\% & 1.66\% & 6.38\% & 1.28\% & 0.49\%\\
&\textbf{Ours} (Avg.)  + CtxIns & \textbf{8.92}\% & \textbf{3.54}\% & \textbf{1.60}\% &\textbf{ 0.83}\% & \textbf{4.34}\% & 12.10\% & 12.75\% & 4.20\%\\
& \textbf{Ours} (Wavg.)  + CtxIns & 7.99\% & 3.22\% & 1.50\% & 0.79\% & 4.21\% & 11.86\% & 12.37\% & \textbf{4.25}\%\\
& \textbf{Ours} (TBB)  + CtxIns & 8.32\% & 3.31\% & 1.52\% & 0.80\% & 4.27\% & \textbf{12.11}\% & 12.70\% & 4.19\%\\
& \textbf{Ours} (Avg.) + AttIns &  8.63\% & 3.45\% & 1.57\% & 0.81\% & 4.23\% & 11.72\% & 12.70\% & 4.20\%\\
& \textbf{Ours} (Wavg.) + AttIns &  7.70\% & 3.13\% & 1.44\% & 0.74\% & 4.11\% & 11.54\% & 12.53\% & \textbf{4.25}\%\\
& \textbf{Ours} (TBB) + AttIns &  8.04\% & 3.23\% & 1.47\% & 0.76\% & 4.17\% & 11.81\% & \textbf{12.79}\% & 4.19\%\\
\midrule
&  \textbf{{$Human^\dagger$}}- (Estimation)  & 14.24\% & 7.70\% & 4.76\% &3.22\% & 10.03\% & 15.98\% & 39.58\% & 13.87\%\\
\bottomrule
\end{tabular}
\begin{tablenotes}
            \item *: Reported results are based on our own implementation.
            \item $^\dagger$: Indicative performance, based on two subjects' captions over a subset of 20 samples.
\end{tablenotes}
\end{threeparttable}
\label{Task2}
\end{table*}

\subsection{Implemention Details}
\label{implementation}
We coded our models in PyTorch. We have used the 5th layer of ResNet-152 \cite{he2016deep} for image attention and a single-layer LSTM with dimension size $512$. We re-sized each image into $256\times256$ and then randomly cropped them to $224\times224$. We created our vocabulary by removing words that occur less than $4$ times, resulting in $35$K words while we also truncated long sentences to a maximum length of $31$ words. 
For the article encoding, we used SpaCy's pretrained GloVe embedding with dimension size of $300$ and set the maximum sentence length to $55$.  In $95\%$ of the cases, articles have less than $55$ sentences. In the case of articles with more than $55$ sentences, we encode the average representation of the rest of the sentences at the $55^{th}$ dimension. In all of our models, we used Adam \cite{kingma2014adam} optimizer with $0.002$ learning rate with learning rate decay $0.8$ after $10$ epochs for every $8$ epochs with dropout probability set to $0.2$. We have produced our captions with beam size $1$. The code and dataset are available online\footnote{\url{https://github.com/furkanbiten/GoodNews}}.

\section{Experiments}
\label{experiments}
In this section we provide several experiments in order to evaluate the quality of the image captions generated with our model on the \textit{GoodNews} dataset. First, we compare the obtained results with the state of the art on image captioning using standard metrics. Then we analyze the precision and recall of our method for the specific task of named entity insertion. Finally we provide a human evaluation study and show some qualitative results. 

As discussed extensively in the literature \cite{devlin2015language, elliott2014comparing, Kilickaya2016, Vinyals2016, cui2018learning} standard evaluation metrics for image captioning have several flaws and in many cases they do not correlate with human judgments. Although we present the results in Bleu \cite{Papineni2002}, METEOR \cite{denkowski2014meteor}, ROUGE \cite{lin2004rouge}, CIDEr \cite{vedantam2015cider} and SPICE \cite{anderson2016spice}, we believe the most suitable metric for the specific scenario of image captioning for news images is CIDEr. This is because both METEOR and SPICE use synonym matching and lemmatization, and named entities rarely have any meaningful synonyms or lemmas. For Bleu and ROUGE, every word alters the metric equally: e.g. missing a stop word has the same impact as the lack of a named entity. That is why we believe CIDEr, although it has its own drawbacks, is the most informative metric to analyze our results since it downplays the stop words and puts more importance to the ``unique'' words by using a tf-idf weighting scheme. 
\begin{figure*}[t]
\setlength\extrarowheight{5pt} 
\begin{tabularx}{\linewidth}{@{\extracolsep{\fill}} llX}
\toprule
\multirow{3}{*}{(a)} & \multirow{3}{*}{\includegraphics[width=0.1\textwidth]{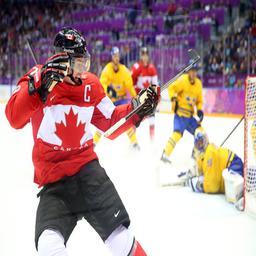}} & \textbf{GT}: \small{\textmd{Sidney Crosby celebrated his goal in the second period that seemed to deflate Sweden.}} \\
& & \textbf{V}: \small{\texttt{Crosby of Vancouver won the Crosby in several seasons.}} \\
& & \textbf{V+T}: \small{\texttt{Crosby of Canada after scoring the winning goal in the second period.}} \\
\midrule
\multirow{3}{*}{(b)} & \multirow{3}{*}{\includegraphics[width=0.1\textwidth]{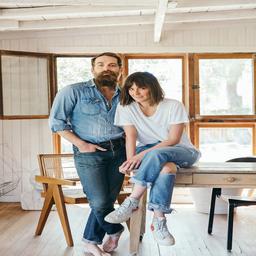}} & \textbf{GT}: \small{\textmd{Ms Ford and her husband Erik Allen Ford in their cabin.}}\\
& & \textbf{V}: \small{\texttt{Leanne Ford and Ford in the kitchen.}} \\
& & \textbf{V+T}: \small{\texttt{Ford and Ford in their home in Echo Park.}} \\
\midrule
\multirow{3}{*}{(c)} & \multirow{3}{*}{\includegraphics[width=0.1\textwidth]{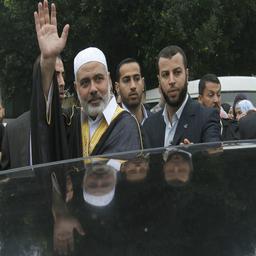}} & \textbf{GT}: \small{\textmd{Ismail Haniya the leader of the Hamas government in Gaza in Gaza City last month.}} \\
& & \textbf{V}: \small{\texttt{Haniya left and Mahmoud Abbas in Gaza City.}} \\
& & \textbf{V+T}:  \small{\texttt{Haniya the Hamas speaker leaving a meeting in Gaza City on Wednesday.}} \\
\midrule
\multirow{3}{*}{(d)} & \multirow{3}{*}{\includegraphics[width=0.1\textwidth]{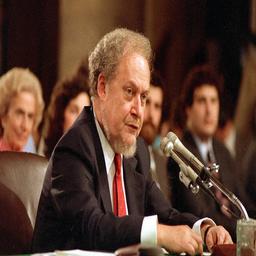}} & \textbf{GT}: \small{\textmd{Supreme Court nominee Robert Bork testifying before the Senate Judiciary Committee.}}\\
& & \textbf{V}: \small{\texttt{Bork left and the Bork Battle in GPE.}} \\
& & \textbf{V+T}:  \small{\texttt{Bork the the Supreme Court director testifying before Senate on 1987.}} \\
\bottomrule
\end{tabularx}
\caption{Qualitative Result; V: Visual Only, V+T: Visual and Textual, GT: Ground Truth}
\label{qualitative}
\end{figure*}
\subsection{News Image Captioning}
Our pipeline for news image captioning operates at two levels. First it produces template captions, before substituting the placeholders with named entities from the text.

Table \ref{FirstNoArticle} shows the results on the intermediate task of template caption generation for state-of-the-art captioning models without using any contextual information (``Visual only'', i.e. ignoring the news articles), and compares them with our method's results using different Article Encoding strategies (``Visual \& Textual''). We appreciate that the ``Show, Attend and Tell''~\cite{Xu2015} model outperforms the rest of the baselines~\cite{anderson2017bottom,rennie2017self,lu2017knowing} on the intermediate task of template caption generation. This outcome differs from the results obtained on other standard benchmarks for image captioning like MSCOCO, where~\cite{anderson2017bottom,rennie2017self,lu2017knowing} are known to improve over the ``Show, Attend and Tell'' model. We believe this discrepancy can be explained because those architectures are  better at recognizing  objects in the input image and their relations, but when the image and its caption are loosely related at the object level, as is the case in the many of the \textit{GoodNews} samples, these models fail to capture the underlying semantic relationships between images and captions. 

Therefore, we have decided to use the architecture of ``Show Attend and Tell'' as the basis for our own model design. We build our two stream architecture, that combines a visual input and a textual input.
From \autoref{FirstArticle}, we can see that encoding the article by simply averaging the GloVe descriptors of its sentences achieves slightly better scores on the intermediate task of template-based captioning than the weighted average and tough-to-beat baseline (TBB) approaches. Overall, the performance of our two-stream (visual and textual) architecture is on par with the baseline results in this task. 

In \autoref{Task2}, we produce the full final captions for both approaches (visual only and visual+textual) by using different strategies for the named entity insertion: random insertion (RandIns), GloVe based context insertion (CtxIns), and insertion by attention (AttIns). Our architecture consistently outperforms the ``Visual only'' pipelines on every metric. Moreover, without the two-stage formulation we introduced (template-based and full captions), current captioning systems (see ``Show Attend Tell - No-NE'' in \autoref{Task2}) as well as BreakingNews \cite{ramisa2018breakingnews} perform rather poorly. 

Despite the fact that the proposed approach yields better results than previous state of the art, and properly deals with out-of-dictionary words (named entities), the overall low results, compared with the typical results on simpler datasets such as MSCOCO, are indicative of the complexity of the problem and the limitations of current captioning approaches. To emphasize this aspect we provide in Table 3 an estimation of human performance in the task of full caption generation on the \textit{GoodNews} dataset. The reported numbers indicate the average performance of $2$ subjects tasked with creating captions for a small subset of $20$ images and their associated articles.

Finally, we provide in \autoref{qualitative} a qualitative comparison for the best performing model of both ``visual only'' (Show, Attend and Tell+CtxIns) and ``visual+textual'' (Avg+AttIns) architectures. We appreciate that taking the textual content into account results in more contextualized captions. We also present some failure cases in which incorrect named entities have been inserted. 

\subsection{Evaluation of Named Entity Insertion} 
Results of \autoref{FirstNoArticle} represent a theoretical maximum, since a perfect named entity insertion would give us those same results for the full caption generation task. However, from \autoref{FirstNoArticle} results to \autoref{Task2} there is a significant drop ranging from $4$ to $18$ points in each metric. To further quantify the differences between context insertion and insertion by attention, we provide in \autoref{table:pr} their precision and recall for exact and partial match named entity insertion. In the exact match evaluation, we only accept the insertion of the names as true positive if they match the ground truth character by character, while on the partial match setting, we do consider token level match as being correct (i.e. ``Falletta'' is considered a true positive for the ``JoAnn Falletta'' entity). 
\begin{table}[t]
\caption{Precision and Recall for named entity insertion.}
\centering
\begin{tabular}{lcccc}
\toprule
       & \multicolumn{2}{c}{{\small Exact match}} & \multicolumn{2}{c}{{\small Partial match}}\\
       & \multicolumn{1}{c}{P} & \multicolumn{1}{c}{R} & \multicolumn{1}{c}{P} & \multicolumn{1}{c}{R}\\
\midrule
Show Attend Tell + CtxIns & 8.19 & 7.10 & 19.39 & 17.33\\
\midrule
Ours (Avg.) + CtxIns & 8.17 & 7.23 & 19.53 & 17.88\\
Ours (WAvg.) + CtxIns & 7.80 & 6.68 & 19.14 & 17.08\\
Ours (TBB) + CtxIns & 7.84 & 6.64 & 19.60 & 17.11\\
Ours (Avg.) + AttIns & \textbf{9.19} & \textbf{8.21} & 21.17 & \textbf{19.48}\\
Ours (WAvg.) + AttIns & 8.88 & 7.74 & 21.11 & 19.00\\
Ours (TBB) + AttIns & 9.09 & 7.81 & \textbf{21.71} & 19.19\\
\bottomrule
\end{tabular}
\label{table:pr}
\vspace{-0.3cm}
\end{table}
\begin{figure}[h]
    \centering
    \includegraphics[width=\linewidth]{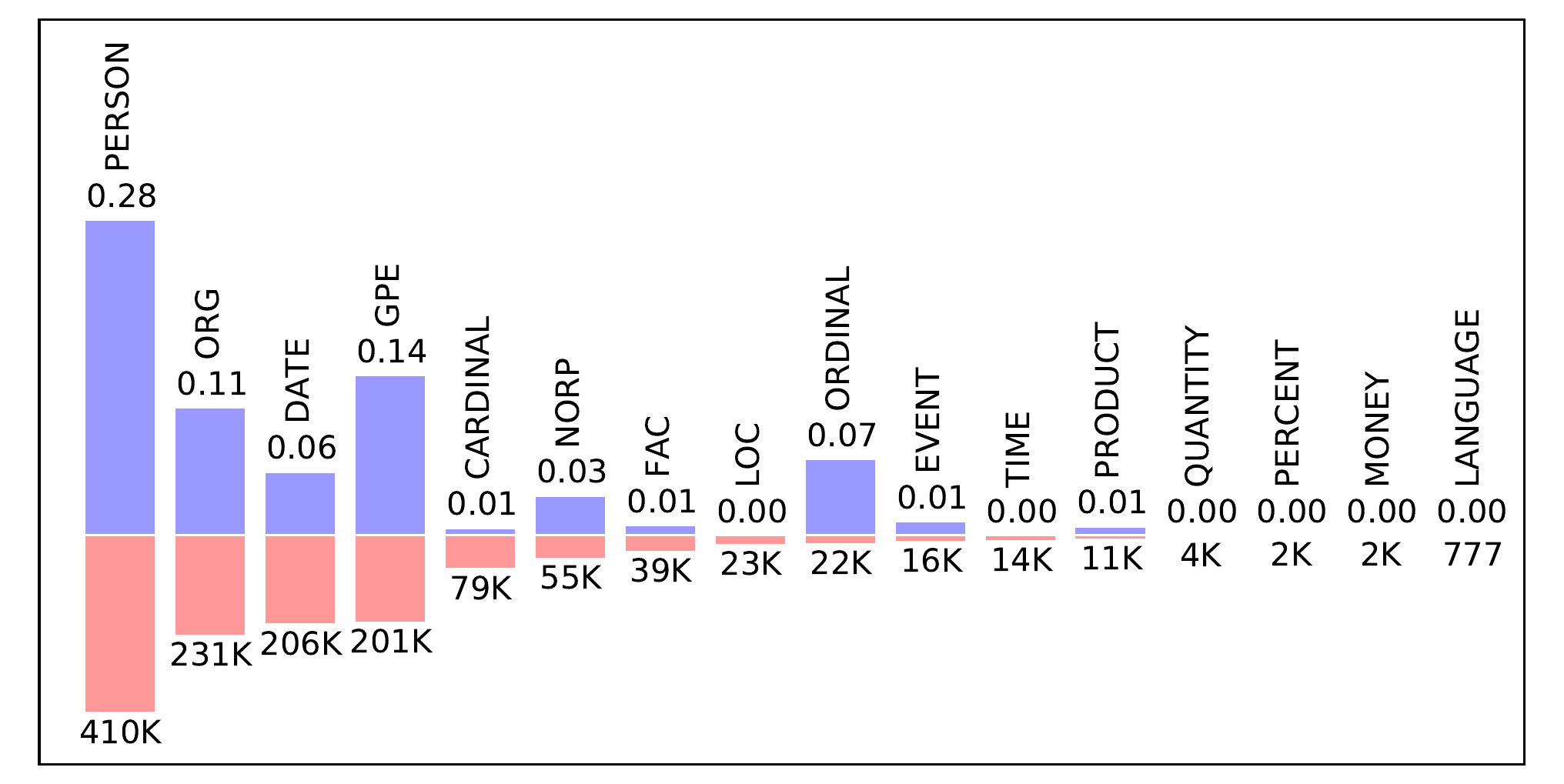}
    \caption{Named entity insertion recall (blue) and number of training samples (red) for each named entity category.}
    \label{fig:recall_per_class}
\end{figure} 
In \autoref{table:pr}, we observe that the proposed insertion by attention (``AttIns'') clearly outperforms the ``CtxIns'' strategy at both exact and partial match evaluations. 
The use of the proposed text attention mechanism allows us to deal with named entity insertion in an end-to-end fashion, eliminating the need for any separate processing.

However, notice that this was not revealed by the analysis of \autoref{Task2}, where all insertion strategies seem to have a similar effect.
This is partly explained by the fact that image captioning evaluation metrics fail to put any special weight to named entities. Intuitively, humans would prefer captions where the named entities are correctly inserted.
To further analyze the results of this experiment we provide in \autoref{fig:recall_per_class} the named entity insertion recall of our method (Avg+AttIns) on each of the individual named entity tags. We observe a correlation of the recall values with the number of training samples for each named entity category. This suggests that the overall named entity insertion performance can be potentially improved with more training data. 





\subsection{Human Evaluation}

In order to provide a more fair evaluation we have conducted a human evaluation study. We asked $20$ human evaluators to compare the outputs of the best performing ``visual + textual'' model (Avg. + AttIns) with the ones of the best performing ``visual only'' model (``Show Attend and Tell'' with Ctx named entity insertion) on a subset of $106$ randomly chosen images. Evaluators were presented an image, its ground-truth caption, and the two captions generated by those methods, and were asked to choose the one they considered most similar to the ground truth. In total we collected $2,101$ responses. 

\begin{figure}[h]
    \centering
    \includegraphics[width=\linewidth]{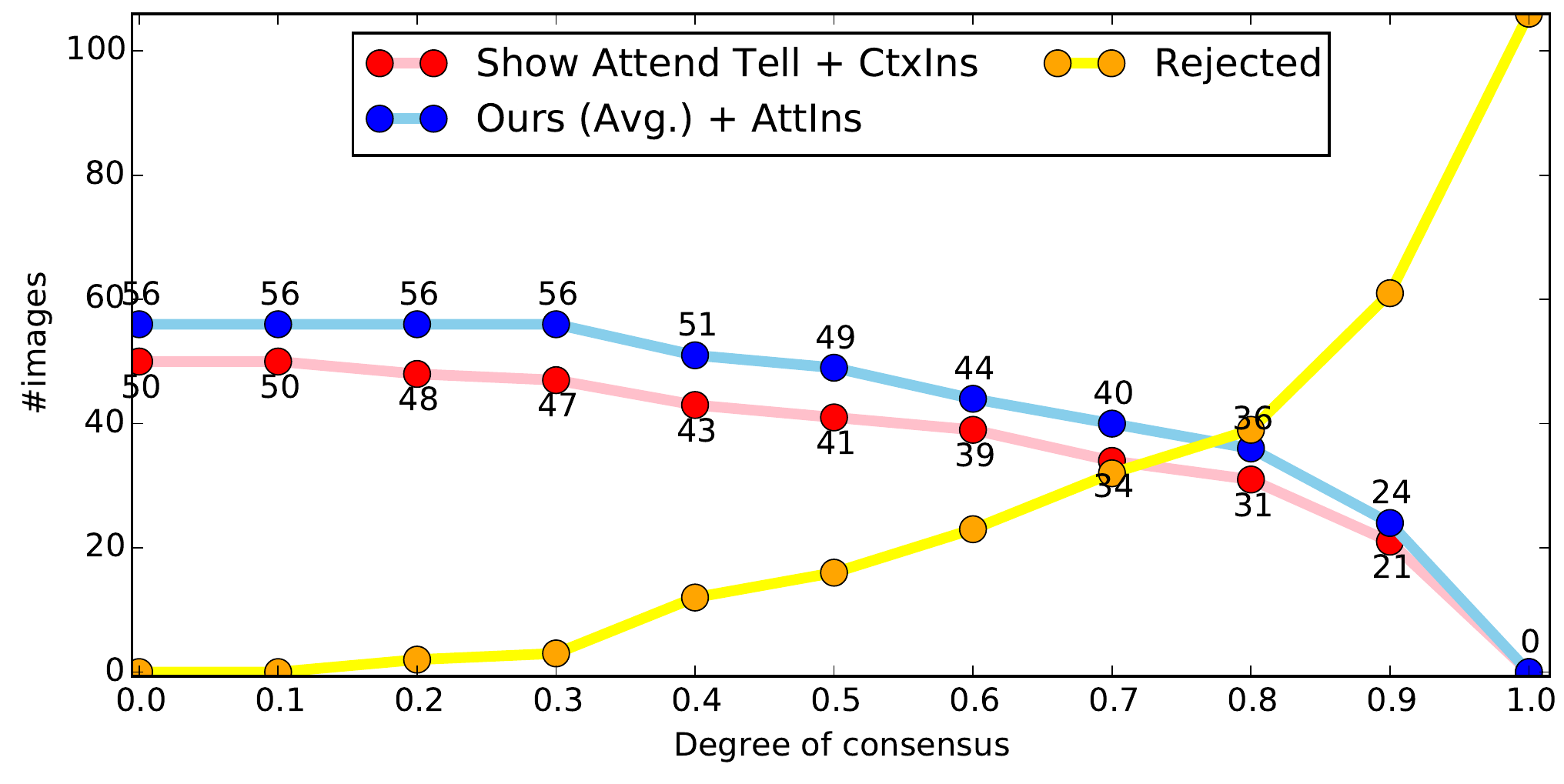}
    \caption{Comparison of ``visual only'' and ``visual+textual'' models regarding human judgments.}
    \label{fig:human_study}
\end{figure}

The comparative study revealed that our model was perceived as better than ``Show Attend and Tell + CtxIns'' in $53\%$ of
the cases. In \autoref{fig:human_study} we analyze the results as a function of the degree of consensus of the evaluators for each image. Our aim is to exclude from the analysis those images in which there is no clear consensus about the better caption between the evaluators. 
To do this we define the degree of consensus $C = 1 - \frac{min(votes_v, votes_v+t)}{max(votes_v, votes_v+t)}$, where $votes_v$ and $votes_{v+t}$ denote the evaluator votes for each method. At each value of $C$ We reject all images that have smaller consensus. Then we report on how many samples the majority vote was for the ``visual'' or ``visual+textual'' method. 
As can be appreciated the results indicate a consistent preference for the ``visual+textual'' variant.

\section{Conclusion}
\label{conclusion}
In this paper we have presented a novel captioning pipeline that aims to take a step closer to producing captions that offer a plausible interpretation of the scene, and applied it to the particular case of news image captioning. In addition, we presented \textit{GoodNews}, a new dataset comprising $466$K samples, the largest news-captioning dataset to date.
Our proposed pipeline integrates contextual information, given here in the form of a news article, introducing an attention mechanism that permits the captioning system to selectively draw information from the context source, guided by the image.
Furthermore, we proposed a two-stage procedure implemented in an end-to-end fashion, to incorporate named entities in the captions, specifically designed to deal with out-of-dictionary entities that are only made available at test time. 
Experimental results demonstrate that the proposed method yields state-of-the-art performance, while it satisfactorily incorporates named entity information in the produced captions. 
\vspace{-0.2cm}

\section*{Acknowledgements}
 \vspace{-0.2cm}

\small{This work has been supported by projects TIN2017-89779-P, Marie-Curie (712949 TECNIOspring PLUS), aBSINTHE (Fundaci\'on BBVA 2017), the CERCA Programme / Generalitat de Catalunya, NVIDIA Corporation and a UAB PhD scholarship.}


{\small

\input{egbib.bbl}
\bibliographystyle{ieee}
}
\end{document}

%% file: model_figure.tex
\pdfoutput=1
\begingroup%
  \makeatletter%
  \providecommand\color[2][]{%
    \errmessage{(Inkscape) Color is used for the text in Inkscape, but the package 'color.sty' is not loaded}%
    \renewcommand\color[2][]{}%
  }%
  \providecommand\transparent[1]{%
    \errmessage{(Inkscape) Transparency is used (non-zero) for the text in Inkscape, but the package 'transparent.sty' is not loaded}%
    \renewcommand\transparent[1]{}%
  }%
  \providecommand\rotatebox[2]{#2}%
  \ifx\svgwidth\undefined%
    \setlength{\unitlength}{743.83362347bp}%
    \ifx\svgscale\undefined%
      \relax%
    \else%
      \setlength{\unitlength}{\unitlength * \real{\svgscale}}%
    \fi%
  \else%
    \setlength{\unitlength}{\svgwidth}%
  \fi%
  \global\let\svgwidth\undefined%
  \global\let\svgscale\undefined%
  \makeatother%
  \begin{picture}(1,0.457)%
    \put(0,0){\includegraphics[width=\unitlength]{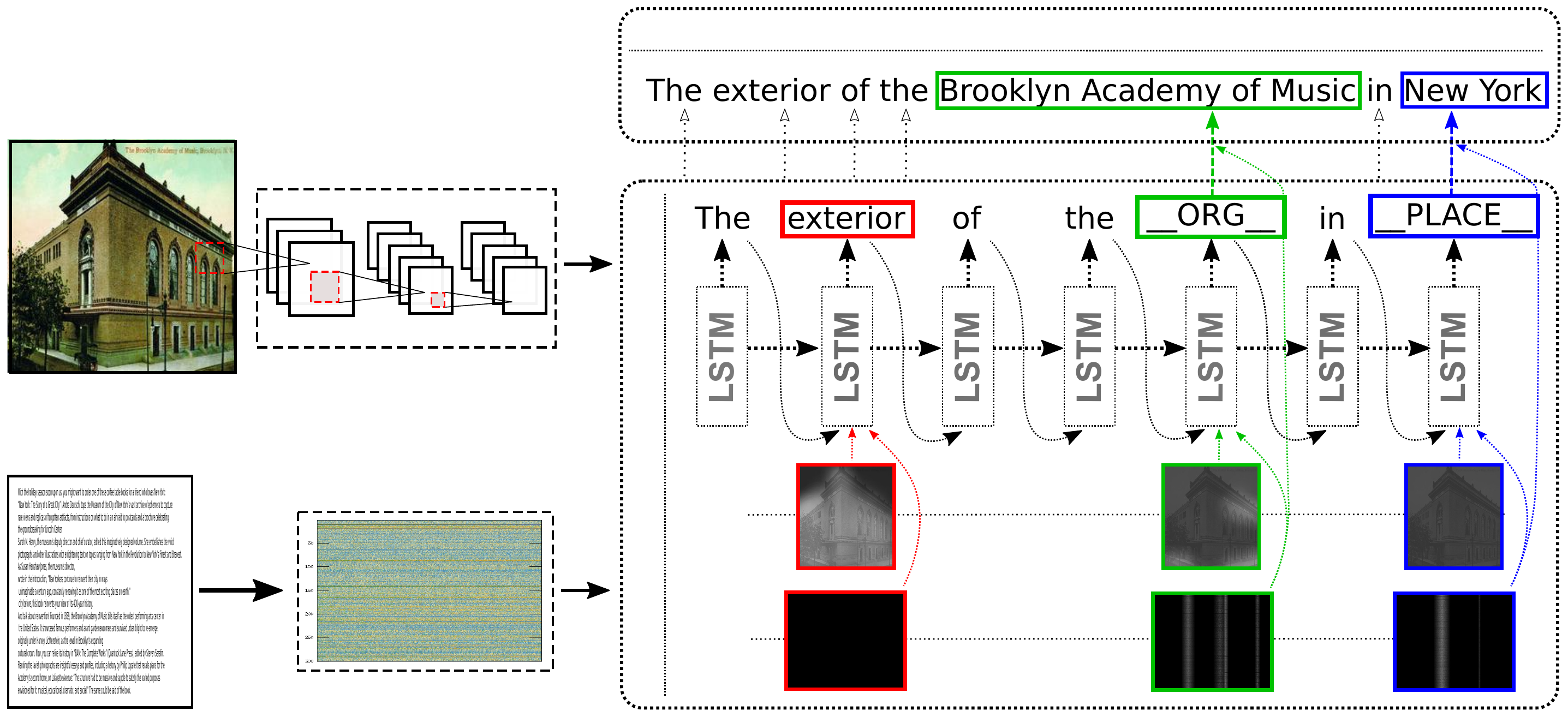}}%
    \put(0.033,0.375){\color[rgb]{0.23,0.23,0.23}\makebox(0,0)[lb]{\smash{\Large \bf{\emph{Input image}}}}}%
    \put(0.165,0.355){\color[rgb]{0.23,0.23,0.23}\makebox(0,0)[lb]{\smash{\Large \bf{\emph{CNN feature extraction}}}}}%
    \put(0.192,0.205){\color[rgb]{0.23,0.23,0.23}\makebox(0,0)[lb]{\smash{\Large \bf{\emph{$(14 \times 14 \times 256)$}}}}}
    \put(0.417,0.100){\color[rgb]{0.23,0.23,0.23}\makebox(0,0)[lb]{\rotatebox[origin=c]{90}{\smash{\Large \bf{\emph{RNN with Attention}}}}}}%
    \put(0.615,0.135){\color[rgb]{0.23,0.23,0.23}\makebox(0,0)[lb]{\smash{\Large \bf{\emph{Visual}}}}}%
    \put(0.603,0.11){\color[rgb]{0.23,0.23,0.23}\makebox(0,0)[lb]{\smash{\Large \bf{\emph{Attention}}}}}%
    \put(0.608,0.055){\color[rgb]{0.23,0.23,0.23}\makebox(0,0)[lb]{\smash{\Large \bf{\emph{Sentence}}}}}%
    \put(0.606,0.03){\color[rgb]{0.23,0.23,0.23}\makebox(0,0)[lb]{\smash{\Large \bf{\emph{Attention}}}}}%
    \put(0.595,0.431){\color[rgb]{0.23,0.23,0.23}\makebox(0,0)[lb]{\smash{\Large \bf{\emph{Named Entity Insertion}}}}}%
    \put(0.025,0.163){\color[rgb]{0.23,0.23,0.23}\makebox(0,0)[lb]{\smash{\Large \bf{\emph{Input text}}}}}%
    \put(0.139,0.146){\color[rgb]{0.23,0.23,0.23}\makebox(0,0)[lb]{\smash{\Large \bf{\emph{Sentence-level Global Vectors}}}}}
    \put(0.142,0.005){\color[rgb]{0.23,0.23,0.23}\makebox(0,0)[lb]{\smash{\Large \bf{\emph{$(GloVe\_dim \times \#\_sentences)$}}}}}
  \end{picture}%
\endgroup%

%% file: ms.bbl
\begin{thebibliography}{10}\itemsep=-1pt

\bibitem{anderson2016spice}
Peter Anderson, Basura Fernando, Mark Johnson, and Stephen Gould.
\newblock Spice: Semantic propositional image caption evaluation.
\newblock In {\em European Conference on Computer Vision}, 2016.

\bibitem{anderson2017bottom}
Peter Anderson, Xiaodong He, Chris Buehler, Damien Teney, Mark Johnson, Stephen
  Gould, and Lei Zhang.
\newblock Bottom-up and top-down attention for image captioning and {VQA}.
\newblock {\em arXiv preprint arXiv:1707.07998}, 2017.

\bibitem{arora2016simple}
Sanjeev Arora, Yingyu Liang, and Tengyu Ma.
\newblock A simple but tough-to-beat baseline for sentence embeddings.
\newblock In {\em International Conference on Learning Representations}, 2017.

\bibitem{bai2018survey}
Shuang Bai and Shan An.
\newblock A survey on automatic image caption generation.
\newblock {\em Neurocomputing}, 311:291--304, 2018.

\bibitem{bernardi2016automatic}
Raffaella Bernardi, Ruket Cakici, Desmond Elliott, Aykut Erdem, Erkut Erdem,
  Nazli Ikizler-Cinbis, Frank Keller, Adrian Muscat, and Barbara Plank.
\newblock Automatic description generation from images: A survey of models,
  datasets, and evaluation measures.
\newblock {\em Journal of Artificial Intelligence Research}, 55:409--442, 2016.

\bibitem{cui2018learning}
Yin Cui, Guandao Yang, Andreas Veit, Xun Huang, and Serge Belongie.
\newblock Learning to evaluate image captioning.
\newblock In {\em IEEE Conference on Computer Vision and Pattern Recognition},
  2018.

\bibitem{denkowski2014meteor}
Michael Denkowski and Alon Lavie.
\newblock Meteor universal: Language specific translation evaluation for any
  target language.
\newblock In {\em Workshop on Statistical Machine Translation}, 2014.

\bibitem{devlin2015language}
Jacob Devlin, Hao Cheng, Hao Fang, Saurabh Gupta, Li Deng, Xiaodong He,
  Geoffrey Zweig, and Margaret Mitchell.
\newblock Language models for image captioning: The quirks and what works.
\newblock {\em arXiv preprint arXiv:1505.01809}, 2015.

\bibitem{devlin2015exploring}
Jacob Devlin, Saurabh Gupta, Ross Girshick, Margaret Mitchell, and C~Lawrence
  Zitnick.
\newblock Exploring nearest neighbor approaches for image captioning.
\newblock {\em arXiv preprint arXiv:1505.04467}, 2015.

\bibitem{elliott2014comparing}
Desmond Elliott and Frank Keller.
\newblock Comparing automatic evaluation measures for image description.
\newblock In {\em Annual Meeting of the Association for Computational
  Linguistics}, 2014.

\bibitem{Fang2015}
Hao Fang, Saurabh Gupta, Forrest Iandola, Rupesh~K. Srivastava, Li Deng, Piotr
  Doll{\'{a}}r, Jianfeng Gao, Xiaodong He, Margaret Mitchell, John~C. Platt,
  C.~Lawrence Zitnick, and Geoffrey Zweig.
\newblock {From captions to visual concepts and back}.
\newblock In {\em IEEE Conference on Computer Vision and Pattern Recognition},
  2015.

\bibitem{feng2013automatic}
Yansong Feng and Mirella Lapata.
\newblock Automatic caption generation for news images.
\newblock {\em IEEE Transactions on Pattern Analysis and Machine Intelligence},
  35(4):797--812, 2013.

\bibitem{gan2017stylenet}
Chuang Gan, Zhe Gan, Xiaodong He, Jianfeng Gao, and Li Deng.
\newblock Stylenet: Generating attractive visual captions with styles.
\newblock In {\em IEEE Conference on Computer Vision and Pattern Recognition},
  2017.

\bibitem{he2016deep}
Kaiming He, Xiangyu Zhang, Shaoqing Ren, and Jian Sun.
\newblock Deep residual learning for image recognition.
\newblock In {\em IEEE Conference on Computer Vision and Pattern Recognition},
  2016.

\bibitem{spacy2}
Matthew Honnibal and Ines Montani.
\newblock {SpaCy} 2: Natural language understanding with bloom embeddings,
  convolutional neural networks and incremental parsing.
\newblock 2017.

\bibitem{Karpathy}
Andrej Karpathy and Li Fei-Fei.
\newblock {Deep Visual-Semantic Alignments for Generating Image Descriptions}.
\newblock {\em IEEE Transactions on Pattern Analysis and Machine Intelligence},
  39(4):664--676, 2017.

\bibitem{Kilickaya2016}
Mert Kilickaya, Aykut Erdem, Nazli Ikizler-Cinbis, and Erkut Erdem.
\newblock {Re-evaluating Automatic Metrics for Image Captioning}.
\newblock 2016.

\bibitem{kingma2014adam}
Diederik~P Kingma and Jimmy Ba.
\newblock Adam: A method for stochastic optimization.
\newblock {\em arXiv preprint arXiv:1412.6980}, 2014.

\bibitem{lake2017building}
Brenden~M Lake, Tomer~D Ullman, Joshua~B Tenenbaum, and Samuel~J Gershman.
\newblock Building machines that learn and think like people.
\newblock {\em Behavioral and Brain Sciences}, 40, 2017.

\bibitem{lin2004rouge}
Chin-Yew Lin.
\newblock Rouge: A package for automatic evaluation of summaries.
\newblock In {\em ACL Workshop on Text Summarization}, 2004.

\bibitem{lin2014microsoft}
Tsung-Yi Lin, Michael Maire, Serge Belongie, James Hays, Pietro Perona, Deva
  Ramanan, Piotr Doll{\'a}r, and C~Lawrence Zitnick.
\newblock Microsoft {COCO}: Common objects in context.
\newblock In {\em European Conference on Computer Vision}, 2014.

\bibitem{lu2018entity}
Di Lu, Spencer Whitehead, Lifu Huang, Heng Ji, and Shih-Fu Chang.
\newblock Entity-aware image caption generation.
\newblock {\em arXiv preprint arXiv:1804.07889}, 2018.

\bibitem{lu2017knowing}
Jiasen Lu, Caiming Xiong, Devi Parikh, and Richard Socher.
\newblock Knowing when to look: Adaptive attention via a visual sentinel for
  image captioning.
\newblock In {\em Conference on Computer Vision and Pattern Recognition}, 2017.

\bibitem{mathews2018semstyle}
Alexander Mathews, Lexing Xie, and Xuming He.
\newblock Semstyle: Learning to generate stylised image captions using
  unaligned text.
\newblock In {\em IEEE Conference on Computer Vision and Pattern Recognition},
  2018.

\bibitem{mathews2016senticap}
Alexander~Patrick Mathews, Lexing Xie, and Xuming He.
\newblock Senticap: Generating image descriptions with sentiments.
\newblock In {\em Conference of the Association for the Advancement of
  Artificial Intelligence}, 2016.

\bibitem{nadeau2007survey}
David Nadeau and Satoshi Sekine.
\newblock A survey of named entity recognition and classification.
\newblock {\em Lingvisticae Investigationes}, 30(1):3--26, 2007.

\bibitem{Papineni2002}
Kishore Papineni, Salim Roukos, Todd Ward, and Wj Zhu.
\newblock {BLEU: a method for automatic evaluation of machine translation}.
\newblock {\em Annual Meeting on Association for Computational Linguistics},
  2002.

\bibitem{pennington2014glove}
Jeffrey Pennington, Richard Socher, and Christopher Manning.
\newblock Glove: Global vectors for word representation.
\newblock In {\em Conference on Empirical Methods in Natural Language
  Processing}, 2014.

\bibitem{plummer2015flickr30k}
Bryan~A Plummer, Liwei Wang, Chris~M Cervantes, Juan~C Caicedo, Julia
  Hockenmaier, and Svetlana Lazebnik.
\newblock Flickr30k entities: Collecting region-to-phrase correspondences for
  richer image-to-sentence models.
\newblock In {\em IEEE International Conference on Computer Vision}, 2015.

\bibitem{ramisa2018breakingnews}
Arnau Ramisa, Fei Yan, Francesc Moreno-Noguer, and Krystian Mikolajczyk.
\newblock Breakingnews: Article annotation by image and text processing.
\newblock {\em IEEE Transactions on Pattern Analysis and Machine Intelligence},
  40(5):1072--1085, 2018.

\bibitem{rennie2017self}
Steven~J Rennie, Etienne Marcheret, Youssef Mroueh, Jarret Ross, and Vaibhava
  Goel.
\newblock Self-critical sequence training for image captioning.
\newblock In {\em Conference on Computer Vision and Pattern Recognition}, 2017.

\bibitem{russakovsky2015imagenet}
Olga Russakovsky, Jia Deng, Hao Su, Jonathan Krause, Sanjeev Satheesh, Sean Ma,
  Zhiheng Huang, Andrej Karpathy, Aditya Khosla, Michael Bernstein, et~al.
\newblock Imagenet large scale visual recognition challenge.
\newblock {\em International Journal of Computer Vision}, 115(3):211--252,
  2015.

\bibitem{tariq2017context}
Amara Tariq and Hassan Foroosh.
\newblock A context-driven extractive framework for generating realistic image
  descriptions.
\newblock {\em IEEE Transactions on Image Processing}, 26(2):619--632, 2017.

\bibitem{terman1916measurement}
Lewis~Madison Terman.
\newblock {\em The measurement of intelligence: An explanation of and a
  complete guide for the use of the Stanford revision and extension of the
  Binet-Simon intelligence scale}.
\newblock Houghton Mifflin, 1916.

\bibitem{tran2016rich}
Kenneth Tran, Xiaodong He, Lei Zhang, Jian Sun, Cornelia Carapcea, Chris
  Thrasher, Chris Buehler, and Chris Sienkiewicz.
\newblock Rich image captioning in the wild.
\newblock In {\em IEEE Conference on Computer Vision and Pattern Recognition
  Workshops}, 2016.

\bibitem{vedantam2015cider}
Ramakrishna Vedantam, C Lawrence~Zitnick, and Devi Parikh.
\newblock Cider: Consensus-based image description evaluation.
\newblock In {\em IEEE Conference on Computer Vision and Pattern Recognition},
  2015.

\bibitem{vinyals2015show}
Oriol Vinyals, Alexander Toshev, Samy Bengio, and Dumitru Erhan.
\newblock Show and tell: A neural image caption generator.
\newblock In {\em IEEE Conference on Computer Vision and Pattern Recognition},
  2015.

\bibitem{Vinyals2016}
Oriol Vinyals, Alexander Toshev, Samy Bengio, and Dumitru Erhan.
\newblock {Show and Tell: Lessons Learned from the 2015 MSCOCO Image Captioning
  Challenge}.
\newblock {\em IEEE Transactions on Pattern Analysis and Machine Intelligence},
  39(4):652--663, 2017.

\bibitem{Xu2015}
Kelvin Xu, Jimmy Ba, Ryan Kiros, Kyunghyun Cho, Aaron Courville, Ruslan
  Salakhutdinov, Richard Zemel, and Yoshua Bengio.
\newblock {Show, Attend and Tell: Neural Image Caption Generation with Visual
  Attention}.
\newblock In {\em International Conference on Machine Learning}, 2015.

\end{thebibliography}
